%% file: iclr2027_conference.tex
\documentclass{article} % For LaTeX2e
\usepackage{iclr2027_conference,times}
\usepackage{microtype}

\input{math_commands.tex}

\usepackage{hyperref}
\usepackage{url}
\usepackage{booktabs}
\usepackage{graphicx}
\usepackage{xcolor}
\usepackage[breakable]{tcolorbox}
\usepackage{algorithm}
\usepackage{algpseudocode}
\usepackage{amsthm}
\usepackage{amssymb}
\usepackage{multirow}
\usepackage{colortbl}
\definecolor{hostrow}{RGB}{234,242,252}

\title{SIPO: Selective-Inference Policy Optimization for Tree-Structured Agentic RL}

\newcommand{\SIPOAuthorBlock}{%
\begin{minipage}[t]{\dimexpr\textwidth-2\tabcolsep\relax}
\centering
{\normalsize\bfseries
\mbox{Zenghuang Fu\textsuperscript{1,2,*}}\quad
\mbox{Ningqi Chen\textsuperscript{3,*}}\quad
\mbox{Mingda Jia\textsuperscript{1,2,*}}\quad
\mbox{Xiaofeng Han\textsuperscript{1,2}}\quad
\mbox{Zhaoyang Li\textsuperscript{4}}\quad
\mbox{Qiuyuan Ai\textsuperscript{4}}\quad
\mbox{Zelong Zheng\textsuperscript{1,2}}\quad
\mbox{Haoyu Wu\textsuperscript{5}}\quad
\mbox{Tianyu Fu\textsuperscript{5}}\quad
\mbox{Chenxu Zhao\textsuperscript{5}}\quad
\mbox{Minghui Wu\textsuperscript{5}}\quad
\mbox{Guannan He\textsuperscript{4,\textdagger}}\quad
\mbox{Changwei Wang\textsuperscript{6,7,\textdagger}}\par}
\medskip
{\normalsize\normalfont
\textsuperscript{1}University of Chinese Academy of Sciences\par
\textsuperscript{2}Institute of Automation, Chinese Academy of Sciences\par
\textsuperscript{3}The University of Hong Kong\par
\textsuperscript{4}Peking University\par
\textsuperscript{5}Mininglamp Technology\par
\textsuperscript{6}Key Laboratory of Computing Power Network and Information Security,
Ministry of Education; Shandong Computer Science Center,
Qilu University of Technology (Shandong Academy of Sciences)\par
\textsuperscript{7}Key Laboratory of Computing Power Internet and Service Computing,
Shandong Fundamental Research Center for Computer Science\par
\medskip
\textsuperscript{*}These authors contributed equally.\par
\textsuperscript{\textdagger}Co-corresponding authors: Guannan He and Changwei Wang.\par}
\end{minipage}%
}
\author{\SIPOAuthorBlock}
\hypersetup{pdfauthor={Zenghuang Fu, Ningqi Chen, Mingda Jia, Xiaofeng Han,
Zhaoyang Li, Qiuyuan Ai, Zelong Zheng, Haoyu Wu, Tianyu Fu, Chenxu Zhao,
Minghui Wu, Guannan He, Changwei Wang}}

\newcommand{\SFCtag}{SFC}
\newcommand{\XBtag}{EXB}
\newcommand{\OSCtag}{OSC}
\newcommand{\SIPO}{SIPO}

\iclrfinalcopy % Non-anonymous version: show authors and affiliations.
\begin{document}

\maketitle
\lhead{} % Suppress the template's automatic publication-status header.

\begin{abstract}
Tree-structured reinforcement learning trains search agents by comparing alternative
continuations and propagating terminal rewards to intermediate decisions. Adaptive
expansion, however, creates a statistical asymmetry: an incumbent is selected using
its own generation statistic, whereas fresh siblings are sampled after selection.
When that statistic is associated with return, branch values can reflect selection
history as well as continuation quality, even for a shared parent. We propose
Selective-Inference Policy Optimization (\SIPO{}), which incorporates this distinction into tree-based credit
estimation. Its scale-free branch criterion keeps generation scores and sibling
penalties on a consistent relative scale; exchangeable branching supplies multiple
fresh continuations from each selected parent; and order-statistic correction adjusts
retained incumbent values using selection rank and the estimated score--outcome
association. These mechanisms preserve the leaf budget and the host policy optimisation
objective. Across seven QA benchmarks using Qwen3-4B, Qwen3-8B, and Qwen2.5-7B,
\SIPO{} achieves the highest reported multi-hop and single-hop averages among the
compared methods. On Qwen3-8B, it improves these averages over AT\textsuperscript{2}PO
by $1.31$ and $1.07$ percentage points, respectively, and ranks first on six of seven
benchmarks. Component ablations evaluate the individual and combined changes, while
early-training paired diagnostics show a selected--fresh value gap alongside a
near-zero fresh--fresh reference. Together, these results support accounting for
selection history when constructing and evaluating search-agent rollouts.
Our code is available at \url{https://github.com/Zenghuang-Fu/SIPO}.
\end{abstract}

\section{Introduction}
\label{sec:intro}

Large language models can act as search agents by alternating between reasoning, query
generation, and the analysis of retrieved documents \citep{cognitivescaffold,coevokg,sesa}.
Successful search requires a sequence of decisions about which evidence to retrieve,
how to use it, and when sufficient information has been collected to answer a question.
Reinforcement learning with verifiable rewards trains these decisions from final-answer
correctness \citep{deepseekr1,searchr1}, but the terminal reward leaves open how credit
should be assigned to intermediate search and reasoning steps. Group-relative
optimisation compares rewards across trajectories for the same prompt
\citep{deepseekmath}, providing a trajectory-level signal that offers limited
resolution for individual decisions. A successful trajectory can contain unnecessary
tool calls, and an unsuccessful one can contain useful early reasoning. Tree-structured
rollouts provide finer comparisons by sharing prefixes and exploring alternative
continuations \citep{at2po,bpo,treerl}. Adaptive methods decide which intermediate
states to expand and propagate terminal outcomes through the resulting tree, with
recent work improving branch placement, rollout allocation, and value aggregation
\citep{aepo,appo,igrpo,patr}. Tree construction consequently shapes both the actions
explored during training and the observations available for estimating their credit.

Our starting point is that branches with a shared parent can have different selection
histories. Consider an incumbent selected for expansion because its realised surprisal
is high. The sampler returns to the parent context and generates a fresh continuation.
The incumbent has passed a filter based on its own generation, whereas the fresh
sibling is sampled after that filter. When the selection statistic is associated with
return, their conditional expected values can differ before any policy update, even
when they share the same question, retrieved evidence, and terminal reward function.
A shared prefix therefore does not by itself make the two sampling histories
symmetric. If their values are propagated through the tree, this distinction can also
affect the credit assigned to preceding decisions. The coupling between branch
selection and evaluation connects tree-based training to post-selection inference
\citep{berk2013valid,taylor2015statistical} and double estimation
\citep{thrun1993issues,hasselt2010double}. It suggests a concrete diagnostic: two fresh
siblings drawn from the same selected parent, under symmetric continuation and
evaluation rules, provide a reference for the difference between a selected incumbent
and a fresh continuation. Their mean contrast can be examined alongside the
selected--fresh contrast to distinguish the roles of shared context and selection
history in branch-value comparisons.

We propose Selective-Inference Policy Optimization (\SIPO{}), which incorporates
selection history into adaptive tree training while preserving the leaf budget and
host policy objective. The design connects three stages of the training process:
calibrating the scores used to allocate expansions, constructing fresh comparisons
at selected parents, and adjusting retained incumbent values before credit propagation.
Score calibration keeps the sibling penalty on a consistent relative scale as the
policy changes. Fresh sampling supplies continuations with symmetric generation
histories, while the value adjustment accounts for the incumbent's selection rank
and the estimated association between its score and outcome. Selection records link
these stages, allowing the correction to use information from tree construction
without additional rollouts for each incumbent. Across seven QA benchmarks with
Qwen3-4B, Qwen3-8B, and Qwen2.5-7B, \SIPO{} improves multi-hop and single-hop
average accuracy over AT\textsuperscript{2}PO on all three backbones. Component
ablations examine the individual and combined changes, and training curves track
accuracy and policy entropy over optimisation. Paired branch diagnostics examine selected--fresh
and fresh--fresh value differences, while empirical calibration checks the rank model
underlying the correction. These evaluations connect the overall performance results
to the branch comparisons that motivate the method.

Our contributions are threefold:
\begin{itemize}
\item \textbf{Scale-free branch criterion.}
We standardise the generation score before applying the sibling penalty, keeping their
relative scale consistent as the policy changes.
\item \textbf{Exchangeable branching.}
We generate multiple fresh continuations per selected parent to provide symmetric
comparisons, retaining incumbents and reallocating expansion slots to preserve the leaf
count.
\item \textbf{Order-statistic correction.}
We model selected-value displacement through rank and the score--outcome association,
then adjust retained incumbent values before tree propagation without additional
rollouts for each incumbent.
\end{itemize}

\section{Related Work}
\label{sec:related}

\subsection{Reinforcement Learning for Search Agents}
\label{sec:rw:search}

Search-agent reinforcement learning uses answer correctness to train reasoning and
tool use \citep{searchr1}. Recent methods refine turn-level supervision: TSPO rewards
the first occurrence of the reference answer \citep{tspo2026}, TIPS uses teacher-based
potential shaping \citep{tips2026}, and CW-GRPO reweights outcome advantages by judged
round contributions \citep{cwgrpo2026}.

Rollout allocation uses entropy, prefix predictions, information gains, or sequential
decisions \citep{aepo,trace,igrpo,infotree,sara,rail}; TSR applies per-turn search
during training \citep{tsr2026}.

\subsection{Tree-Structured Rollouts and Credit Estimation}
\label{sec:rw:trees}

Tree rollouts share prefixes and compare alternative continuations
\citep{at2po,bpo,treerl,egb,patr}, with branch placement guided by uncertainty or
subsequent continuations \citep{appo}. Symmetric sampling \citep{bpo} provides a
reference for analysing score-selected incumbents.

Other approaches refine credit through hindsight critics (HCAPO;
\citealt{hcapo2026}), grouped temporal advantages (GAGPO; \citealt{gagpo2026}),
state-transition graphs (G2PO; \citealt{g2po2026}), or behavioural groups with
action-conditioned baselines (BiPACE; \citealt{bipace2026}). \SIPO{} tracks branch
selection history when estimating values.

Post-selection inference studies evaluation after selection
\citep{berk2013valid,taylor2015statistical}; double estimation separates selection and
evaluation to address maximisation bias \citep{thrun1993issues,hasselt2010double}.
Our rank model addresses branch-value displacement; cluster-sampling theory
\citep{kish1965,cochran1977} informs the shared-prefix analysis in
Appendix~\ref{app:correlation}.

\section{Preliminaries}
\label{sec:preliminaries}

\paragraph{Search-agent reinforcement learning.}
Let $x$ be a question and $\pi_\theta$ a language-model policy. A search trajectory is
$u=(x,y_1,o_1,\ldots,y_{T-1},o_{T-1},y_T)$, where $y_t$ is a reasoning segment followed
by a tool call or final answer, and $o_t$ is the observation returned by the search tool.
The terminal reward $R(u)$ evaluates the final answer. Training maximises
$J(\theta)=\mathbb{E}_{x\sim\mathcal D,u\sim\pi_\theta(\cdot\mid x)}[R(u)]$.
In a rollout tree $\mathcal T(x)$, a node stores a generated segment and its preceding
context; complete root-to-leaf paths are trajectories. We write $p(v)$ for a node's
parent and $\operatorname{ch}(v)$ for its children. Descendant rewards are aggregated
into node values, which provide credit for the corresponding generated segments.

\paragraph{Adaptive tree rollouts.}
Adaptive tree sampling for search agents \citep{at2po} starts
with $M$ trajectories from the input question. At each of $L$ expansion iterations, a
selection rule chooses $K$ candidate nodes and generates $B$ fresh siblings for each
chosen incumbent. When all expansion slots are filled, the tree contains
\begin{equation}
N=M+LKB
\label{eq:budget}
\end{equation}
leaves. Concrete rollout configurations are specified in Appendix~\ref{app:setup}.
Candidates are ranked by realised surprisal with a penalty proportional to the number
of existing siblings. A selected segment is rolled back, and the new continuation is
generated from its parent context. This procedure enables targeted exploration while
reusing the preceding reasoning and retrieved information.

\section{Method}
\label{sec:method}

\subsection{Overview}
\label{sec:overview}

\begin{figure}[!t]
\centering
\includegraphics[width=\textwidth]{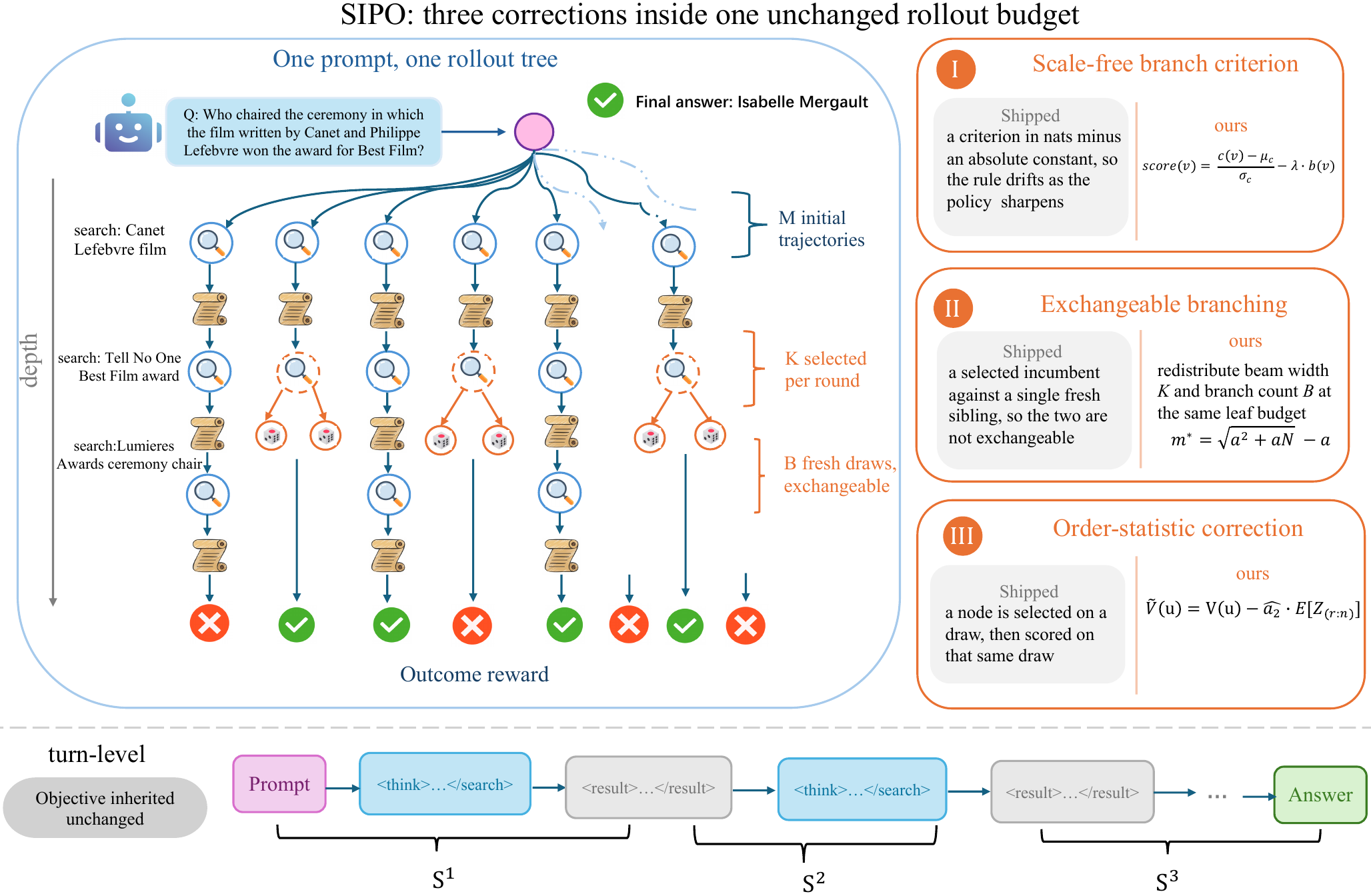}
\caption{Overview of \SIPO{}. The illustrative search tree combines score calibration,
fresh sibling generation, and correction of retained incumbent values. Fresh siblings
have symmetric sampling histories conditional on their parent. The bottom strip shows
the sequence of generated turns and retrieved observations; observations are masked
from the policy loss. The method preserves the leaf budget and the host objective.
The allocation expression in panel II is the simplified surrogate discussed in
Appendix~\ref{app:correlation}; the implemented rollout budget follows Eq.~\ref{eq:budget}.}
\label{fig:framework}
\end{figure}

\SIPO{} coordinates three stages of adaptive tree training (Figure~\ref{fig:framework}).
The \emph{scale-free branch criterion} normalises candidate scores before applying the
sibling penalty. \emph{Exchangeable branching} draws multiple fresh continuations per
selected parent while retaining its incumbent. The \emph{order-statistic correction}
adjusts eligible leaf values using selection records before propagating them into node
advantages. The leaf budget and host policy objective are preserved.

Each expansion also produces a selection record linking the chosen incumbent to its
parent, rank, candidate pool, and fresh siblings. This record connects tree
construction to credit estimation: after terminal rewards become available, it
identifies which branch values carry a selection history. The three components
therefore operate on the same tree at different stages of the training step.

Algorithm~\ref{alg:sipo} in Appendix~\ref{app:algorithm} gives the complete training
loop. The clipped turn-level objective and tool-observation masking are detailed in
Appendix~\ref{app:policy}.

\subsection{Scale-Free Branch Criterion}
\label{sec:sfc}

Adaptive expansion ranks candidates by a generation score minus a sibling-count penalty.
As the policy changes, score-scale drift changes the penalty's relative influence.
When the score spread grows, a fixed penalty becomes weaker relative to score
differences; when the spread contracts, the penalty can dominate the ranking. The
allocation between revisiting a parent and exploring other locations can therefore
drift even when the penalty coefficient is unchanged.
At iteration $\ell$, let $\mathcal C_\ell$ be the candidate set, $s(v)$ the realised
surprisal of candidate $v$, and $b(v)$ its number of existing siblings. Replacing the
host's score $q(v)=s(v)-\lambda b(v)$, \SIPO{} uses
\begin{equation}
\begin{aligned}
q_{\SFCtag}(v)&=\frac{s(v)-\mu_{\mathcal C_\ell}}
{\max(\sigma_{\mathcal C_\ell},\epsilon)}-\lambda b(v),\\
\mathcal S_\ell&=\operatorname{top\text{-}}K(q_{\SFCtag},\mathcal C_\ell).
\end{aligned}
\label{eq:sfc}
\end{equation}
where $\mu_{\mathcal C_\ell}$ and $\sigma_{\mathcal C_\ell}$ are the candidate mean and
standard deviation. This expresses $\lambda$ in standard-deviation units and stabilises
the trade-off between surprisal and sibling count. Calibration changes which parents
receive expansions while preserving the expansion budget $KB$.

\paragraph{Effect on branch allocation.}
Let $d_\ell=\max(\sigma_{\mathcal C_\ell},\epsilon)$. Comparing candidates $v$ and
$v'$ in Eq.~\ref{eq:sfc}, the first ranks higher precisely when
$s(v)-s(v')>\lambda d_\ell[b(v)-b(v')]$. A candidate with more existing siblings
must therefore offer a larger surprisal difference, measured on the current score
scale, to receive another expansion. Mean subtraction cancels in this comparison;
the standard deviation determines the effective penalty in raw-score units.

When sibling counts are equal, normalisation preserves the surprisal ordering.
When the variance floor remains inactive, a positive affine rescaling of all candidate
scores also leaves the calibrated ranking unchanged. Recomputing the statistics
at each expansion makes this trade-off depend on the current candidate population,
using only scores already available from generation.

\subsection{Exchangeable Branching}
\label{sec:exb}

The selected incumbent's own score helped choose its parent for expansion. Fresh
continuations are sampled after that choice, so a shared context alone does not make
them statistically comparable to the incumbent. We generate multiple fresh siblings
from the same parent context and policy, providing a subset with the same conditional
generation distribution. The incumbent remains available for learning.

\paragraph{Sampling from the parent state.}
For each selected incumbent, we restore its parent context, including the question,
preceding generated segments, and retrieved observations. Fresh siblings begin at
this common boundary and sample new continuations under the same policy and remaining
tool budget. They can consequently pursue different queries while reusing the same
evidence. Retaining the incumbent preserves the trajectory already generated;
the new siblings add local alternatives whose labels do not depend on their success.

Fresh labels are assigned independently of realised outcomes. Under symmetric
continuation and evaluation, swapping those labels leaves the joint distribution
unchanged, so their expected advantage difference vanishes. Individual rewards can
still differ. The incumbent has passed the score-based selection filter and does
not share this symmetry with the fresh subset.
Because the parent is fixed before fresh sampling, its selection affects both fresh
labels through the same conditioning context. The reference comparison uses matched
continuation rules, tool limits, and evaluation, so that this symmetry is preserved
for the outcomes being compared.

\paragraph{Allocation under a fixed budget.}
Under the fixed leaf budget in Eq.~\ref{eq:budget}, increasing $B$ is offset by reducing
$K$ so that $KB$ remains constant. This trades the number of expansion locations for
more fresh comparisons at each location; configurations appear in Appendix~\ref{app:setup}.
This allocation concentrates evidence at fewer parent states: local comparisons become
richer, while fewer distinct locations receive expansions. The calibrated criterion
determines where this concentrated sampling is spent.
We use the fresh--fresh reference to assess selected--fresh value displacement;
Appendix~\ref{app:diagnostics} defines both contrasts and their conditions. Fresh
sampling leaves selected incumbent values in the tree, motivating the correction below.

\subsection{Order-Statistic Correction}
\label{sec:osc}

\OSCtag{} models the displacement associated with selecting an incumbent by its own
score and adjusts its value before tree-based credit propagation.

\paragraph{Rank model.}
Consider independent identically distributed candidate pairs $(Z_i,Y_i)$ with
$Z_i\sim\mathcal N(0,1)$ and $\mathbb E[Y_i\mid Z_i]=a_0+a_2Z_i$.
Selecting the $r$-th largest score among $n$ candidates gives
\begin{equation}
\mathbb E[Y_{[r]}-Y_{\mathrm{fresh}}]
=a_2\,\mathbb E[Z_{(r:n)}]\approx a_2\,h(r,n).
\label{eq:osc}
\end{equation}
where the independent fresh outcome has mean $a_0$. The function $h(r,n)$ is Blom's
approximation to the expected standard-normal order statistic \citep{blom1958};
Appendix~\ref{app:proofs} gives its numerical form and derivation. The model links
value displacement to rank, candidate count, and the score--outcome association.
When $a_2$ is negative, highly ranked values are displaced downwards. We use this as
an approximation for rollout candidates with shared prefixes and heterogeneous parents.

The relation follows by conditioning on the candidate scores: the selected outcome
has conditional mean $a_0+a_2Z_{(r:n)}$. Averaging over score realisations and
subtracting the fresh mean cancels $a_0$. The remaining term captures the part of
the outcome difference associated with selecting a particular score rank.

At a fixed descending rank, a larger candidate pool increases the expected selected
score in this model. The rank term describes selection strength, while $a_2$ converts
that strength into outcome units. This separates how strongly a branch was selected
from how informative its score is about return.

\paragraph{Selected-value adjustment.}
We first express terminal rewards in a common scale within each prompt's tree:
\begin{equation}
V_u=\frac{R(u)-\mu_{\mathcal T}}{\sigma_{\mathcal T}+\epsilon}.
\label{eq:leafnorm}
\end{equation}
Here $\mu_{\mathcal T}$ and $\sigma_{\mathcal T}$ are the mean and standard deviation
of the tree's terminal rewards. For each leaf $u$, let $e(u)$ be its most recent applicable
selection event, with rank $r_e$ and candidate count $n_e$. Only actual selections
qualify. A fresh-branch boundary blocks inheritance of an earlier incumbent's event;
a later selection of that fresh branch remains applicable. We compute
\begin{equation}
\widetilde V_u=
\begin{cases}
V_u-w\widehat a_{2,t-1}h(r_e,n_e),&e(u)\ne\varnothing,\\
V_u,&e(u)=\varnothing.
\end{cases}
\label{eq:oscapply}
\end{equation}
where $w$ controls adjustment strength. The slope $\widehat a_{2,t-1}$ uses unadjusted
candidate outcomes from the preceding batch; the first batch uses no correction.
The preceding-batch estimate supplies a shared correction without resampling each
incumbent solely to estimate its value. Event records determine which leaves receive
that correction, preventing an incumbent's rank from being assigned indiscriminately
to fresh descendants sharing its prefix.
With positive adjustment strength and rank term, a negative estimated slope
raises the selected value, whereas a positive slope lowers it. Thus the correction
depends on the observed association between selection score and outcome.

\paragraph{Coefficient estimation and propagation.}
For standardised criterion $Z$ and outcome $Y$ in the value units being corrected,
we estimate $a_2=\operatorname{Cov}(Z,Y)/\operatorname{Var}(Z)$ from the full
selection-time candidate sets, including unselected candidates. The outcome scale is
retained because subtree means need not have unit variance. Appendix~\ref{app:estimators}
specifies the estimator and event handling. Fitting only selected candidates would
condition this association on the selection event itself. Using unadjusted outcomes
also keeps the regression target separate from the correction being estimated. The
new estimate is saved for the next training step.

Corrected leaves are propagated inward using the host's child-softmax weights $w_c$:
\begin{equation}
\widetilde V_n=\sum_{c\in\operatorname{ch}(n)}w_c\widetilde V_c,
\qquad A(n)=\widetilde V_n.
\label{eq:nodeval}
\end{equation}
Each segment's generated tokens receive its node advantage; retrieved observations
remain context and are masked from the loss. At shared ancestors, fresh and adjusted
incumbent branches jointly influence the credit assigned to preceding decisions.
The resulting advantages enter the clipped policy objective in
Appendix~\ref{app:policy}, completing the connection from selection records to the
policy update.

\section{Experiments}
\label{sec:exp}

\subsection{Experimental Setup}
\label{sec:setup}

Multi-hop QA comprises HotpotQA, 2WikiMultihopQA, Musique, and Bamboogle; single-hop
QA comprises NQ, TriviaQA, and PopQA. Multi-hop training uses the HotpotQA-derived
split, and single-hop training uses its corresponding split. We evaluate Qwen3-4B,
Qwen3-8B, and Qwen2.5-7B against ReAct, GRPO, DAPO, GSPO, AEPO, Tree-GRPO where
available, and AT\textsuperscript{2}PO (Table~\ref{tab:published}). The latter is the
primary comparison for the tree-training pipeline; branch diagnostics use the local
host implementation.

All experiments are conducted using eight NVIDIA A800 80GB GPUs.
Training and evaluation use an e5-base-v2 retriever over wiki-18. Host and \SIPO{}
configurations share $M$, $L$, and $N$, with fixed $KB$; Table~\ref{tab:hyperparameters}
and Appendix~\ref{app:setup} give the hyperparameters and rollout configurations.
We report exact-match accuracy (\%). Avg. is the mean over all evaluation questions
within each task family, and we report the checkpoint with the highest Avg. for both
main results and ablations. Evaluation uses greedy decoding without a repetition
penalty. Reported gains are absolute percentage-point differences; branch diagnostics
use tree-bootstrap intervals for within-run uncertainty (Appendix~\ref{app:protocol}).

\subsection{Main Results}
\label{sec:main}

\begin{table}[!t]
\centering
\small
\setlength{\tabcolsep}{4pt}
\caption{Answer accuracy (\%) on multi-hop and single-hop QA benchmarks, evaluated
using exact match. Bold indicates the best score within each backbone, including ties.}
\label{tab:published}
\begin{tabular}{lccccccccc}
\toprule
\multirow{2}{*}{\textbf{Method}} & \multicolumn{5}{c}{\textbf{Multi-Hop QA}} & \multicolumn{4}{c}{\textbf{Single-Hop QA}} \\
\cmidrule(lr){2-6} \cmidrule(lr){7-10}
 & \textbf{Hotpot} & \textbf{2wiki} & \textbf{Musiq} & \textbf{Bamb} & \textbf{Avg.} & \textbf{NQ} & \textbf{TriviaQA} & \textbf{PopQA} & \textbf{Avg.} \\
\midrule
\rowcolor{hostrow} \multicolumn{10}{c}{Backbone Model: Qwen3-4B} \\
ReAct                                     & 30.42 & 32.92 & 12.83 & 44.80 & 30.01 & 26.75 & 53.53 & 35.34 & 41.31 \\
+ GRPO                                    & 44.76 & 51.40 & 21.60 & 50.40 & 46.02 & 45.98 & 65.17 & 49.18 & 54.97 \\
+ DAPO                                    & 45.95 & 51.81 & 21.68 & 51.20 & 46.65 & 47.50 & 65.84 & 51.03 & 56.33 \\
+ GSPO                                    & 47.07 & 49.25 & 22.68 & 50.40 & 45.69 & 46.01 & 64.24 & 48.50 & 54.28 \\
+ AEPO                                    & 46.36 & 51.78 & 23.47 & 50.40 & 46.95 & 45.71 & 64.66 & 50.13 & 55.20 \\
+ AT\textsuperscript{2}PO                 & 49.44 & 52.99 & 24.80 & \textbf{56.80} & 48.81 & \textbf{47.90} & 65.32 & \textbf{51.81} & 56.44 \\
\textbf{+ \SIPO{}}                        &\textbf{50.72}      &\textbf{54.67}       & \textbf{25.65}      &55.20       &\textbf{50.26}       & 47.60 & \textbf{67.70} & 50.80 & \textbf{56.95} \\
\midrule
\rowcolor{hostrow} \multicolumn{10}{c}{Backbone Model: Qwen3-8B} \\
ReAct                                     & 20.66 & 19.05 &  9.56 & 37.60 & 18.66 & 21.16 & 41.81 & 27.37 & 32.19 \\
+ GRPO                                    & 47.01 & 53.69 & 21.35 & 54.40 & 48.03 & 45.70 & 67.42 & 50.17 & 56.29 \\
+ DAPO                                    & 49.64 & 53.91 & 24.05 & 56.00 & 49.40 & 51.99 & 69.02 & 51.90 & 58.53 \\
+ GSPO                                    & 49.59 & 52.55 & 24.35 &  54.4 & 48.56 & 45.56 & 67.75 & 49.66 & 56.15 \\
+ AEPO                                    & 49.17 & 52.97 & 24.01 & 54.40 & 48.62 & 49.92 & 68.31 & 51.77 & 57.94 \\
+ AT\textsuperscript{2}PO                 & \textbf{51.37} & 53.97 & 26.51 & 56.00 & 50.15 & 51.33 & 69.51 & 52.26 & 58.82 \\
\textbf{+ \SIPO{}}                        &51.33       &\textbf{56.11}       &\textbf{27.31}       &\textbf{57.60}        &\textbf{51.46}      & \textbf{53.58} & \textbf{70.06}& \textbf{53.42} & \textbf{59.89} \\
\midrule
\rowcolor{hostrow} \multicolumn{10}{c}{Backbone Model: Qwen2.5-7B} \\
ReAct                                     &  2.85 &  1.94 &  0.58 &  4.00 &  2.10 &  4.34 & 10.67 &  9.32 &  9.23 \\
+ GRPO                                    & 47.94 & 46.89 & 21.27 & 47.20 & 44.48 & 45.56 & 64.86 & 49.92 & 55.20 \\
+ DAPO                                    & 47.50 & 47.93 & 21.27 & 44.00 & 44.91 & 52.24 & 65.00 & 50.01 & 56.08 \\
+ GSPO                                    & 47.35 & 47.30 & 20.32 & 44.00 & 44.40 & 49.64 & 62.87 & 49.75 & 54.81 \\
+ Tree-GRPO                               & 42.39 & 42.01 & 20.15 & 42.40 & 39.79 & 47.56 & 62.69 & 44.75 & 52.04 \\
+ AEPO                                    & 47.05 & 47.53 & 21.03 & 44.00 & 44.51 & 49.00 & 64.13 & 50.21 & 55.45 \\
+ AT\textsuperscript{2}PO                 & \textbf{49.58} & 48.04 & 22.56 & \textbf{51.20} & 45.83 & 52.91 & 64.90 & 50.44 & 56.34 \\
\textbf{+ \SIPO{}}                        & 48.40 & \textbf{49.90} & \textbf{23.80} & 44.80 & \textbf{46.58} & \textbf{53.38} & \textbf{66.68} & \textbf{51.30} & \textbf{57.52} \\
\bottomrule
\end{tabular}
\end{table}

Table~\ref{tab:published} compares \SIPO{} with the baseline methods across seven
QA benchmarks. \SIPO{} achieves the highest multi-hop and single-hop Avg. on all
three backbones, covering all six backbone--task-family comparisons.
AT\textsuperscript{2}PO is the strongest baseline in each of these aggregate
comparisons, making it the primary reference for the gains. Relative to this baseline,
\SIPO{} improves multi-hop Avg. by $1.45$, $1.31$, and $0.75$ points on Qwen3-4B,
Qwen3-8B, and Qwen2.5-7B, respectively, and single-hop Avg. by $0.51$, $1.07$, and
$1.18$ points in the same order. Qwen3-8B reaches the highest absolute averages,
with $51.46$\% on multi-hop and $59.89$\% on single-hop QA; Qwen3-4B reaches
$50.26$\% and $56.95$\%, while Qwen2.5-7B reaches $46.58$\% and $57.52$\%.
The improvements therefore extend across both evaluated Qwen3 model sizes and the
Qwen2.5 backbone, and across both task families. The tree-training comparison uses
the same nominal leaf budget, supporting the effectiveness of \SIPO{}'s combined
branch-selection, sampling, and credit-estimation design under this constraint.
Measured training costs for Qwen2.5-7B multi-hop QA appear in Table~\ref{tab:budget}.

\begin{figure}[!t]
\centering
\includegraphics[width=\textwidth]{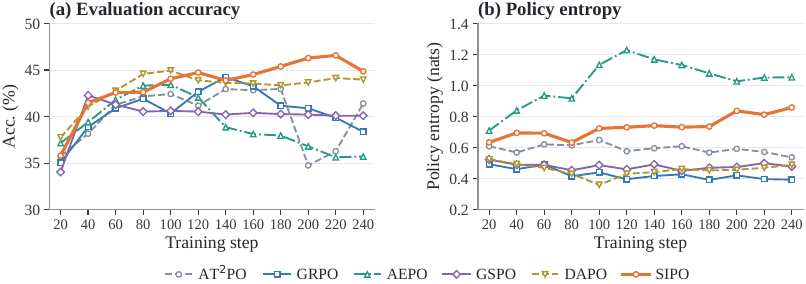}
\caption{Training trajectories on Qwen2.5-7B multi-hop QA for
AT\textsuperscript{2}PO, GRPO, AEPO, GSPO, DAPO, and \SIPO{}.
(a) Size-weighted exact-match accuracy evaluated every 20 training steps.
(b) Policy token entropy recorded at the same steps, without temporal averaging or
smoothing. Each curve corresponds to one completed local run; colors
identify the same method in both panels.}
\label{fig:training-comparison}
\end{figure}

The dataset-level results show where these aggregate gains arise. Relative to
AT\textsuperscript{2}PO, all three backbones improve on 2WikiMultihopQA and Musique,
with gains of $1.68$--$2.14$ and $0.80$--$1.24$ points, respectively, providing a
recurring pattern within the multi-hop evaluations. Qwen3-8B achieves the best score
on six of the seven benchmarks: its improvements include $2.14$ points on
2WikiMultihopQA and $2.25$ points on NQ, while its HotpotQA score is only $0.04$
points below AT\textsuperscript{2}PO. On single-hop QA, Qwen3-8B and Qwen2.5-7B
lead on all three datasets, so their average improvements are accompanied by gains
on NQ, TriviaQA, and PopQA individually. Qwen3-4B exhibits a different distribution:
its $2.38$-point improvement on TriviaQA drives the single-hop average gain despite
lower scores on NQ and PopQA. Multi-hop performance also varies by backbone,
with lower Bamboogle scores for Qwen3-4B and Qwen2.5-7B and a lower HotpotQA score
for Qwen2.5-7B. This breakdown complements the question-weighted averages by
identifying both the recurring improvements and the dataset-specific differences.
Figure~\ref{fig:training-comparison} shows six local Qwen2.5-7B multi-hop training
runs. \SIPO{} attains the highest final accuracy, exceeding DAPO by $0.89$ points.
It also retains higher final policy entropy than AT\textsuperscript{2}PO, GRPO,
GSPO, and DAPO; AEPO has higher entropy but lower final accuracy.
The following ablations examine how the three components contribute to the overall
performance.

\subsection{Ablation Study}
\label{sec:ablation}

\begin{table}[!t]
\centering
\small
\caption{Component ablation on Qwen2.5-7B multi-hop QA. Base denotes the reference
configuration with all three \SIPO{} components disabled.}
\label{tab:ablation}
\begin{tabular}{lcccr}
\toprule
Configuration & \SFCtag & \XBtag & \OSCtag & Avg. (\%)\\
\midrule
Base & -- & -- & -- & 42.98\\
+ \SFCtag & $\checkmark$ & -- & -- & 44.43\\
+ \XBtag & -- & $\checkmark$ & -- & 43.39\\
+ \OSCtag & -- & -- & $\checkmark$ & 43.36\\
+ \SFCtag{} + \XBtag & $\checkmark$ & $\checkmark$ & -- & 44.78\\
\SIPO{} & $\checkmark$ & $\checkmark$ & $\checkmark$ & \textbf{46.58}\\
\bottomrule
\end{tabular}
\end{table}

\begin{figure}[!t]
\centering
\includegraphics[width=\textwidth]{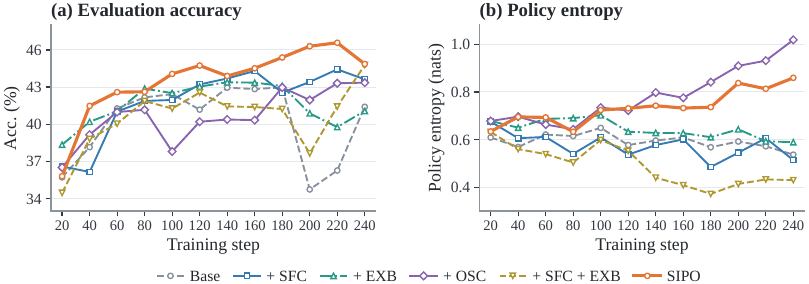}
\caption{Component ablation on Qwen2.5-7B multi-hop QA. (a) Size-weighted accuracy
evaluated every 20 training steps. (b) Policy token entropy sampled at the same
steps, without temporal averaging or smoothing. Each curve represents one ablation
run; colors identify the same configurations in both panels. Base denotes the reference
configuration with all three \SIPO{} components disabled.}
\label{fig:ablation}
\end{figure}

We assess the three components on Qwen2.5-7B multi-hop QA under the same leaf budget
and evaluation protocol (Table~\ref{tab:ablation}). Base disables all three components;
we enable \SFCtag{}, \XBtag{}, and \OSCtag{} individually, combine \SFCtag{} with
\XBtag{}, and then evaluate full \SIPO{}. From Base's $42.98$\%, the individual
components improve Avg. by $1.45$, $0.41$, and $0.38$ points, respectively.
Combining \SFCtag{} and \XBtag{} reaches $44.78$\%; adding \OSCtag{} raises this to
$46.58$\%, a further $1.80$ points and a total gain of $3.60$ points over Base.
In this comparison, \OSCtag{} provides a larger accuracy gain with \SFCtag{} and
\XBtag{} than when enabled alone.
Figure~\ref{fig:ablation} complements these checkpoint summaries with size-weighted
accuracy and policy entropy at matching checkpoints every 20 steps, showing the
training trajectories for all six configurations. Branch-width configurations appear
in Appendix~\ref{app:ablation}; shared-prefix dependence and measured costs are
discussed in Appendix~\ref{app:correlation} and Table~\ref{tab:budget}.

\subsection{Selection Diagnostics}
\label{sec:diagnostic}

\begin{figure}[!t]
\centering
\includegraphics[width=\textwidth]{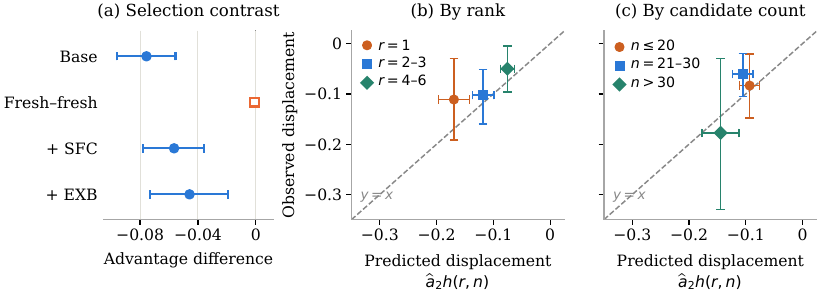}
\caption{Selection displacement and empirical calibration of the rank model.
(a) Early-training selected--fresh node-advantage differences for Base,
$+$\SFCtag{}, and $+$\XBtag{}, with a fresh--fresh reference.
(b)--(c) Predicted versus observed incumbent--fresh displacement on recorded
Qwen2.5-7B multi-hop Base trees, grouped by rank and candidate count.
Panel (a) uses advantage units and separate diagnostic configurations; (b)--(c)
use raw-reward units and the same 1,536 pairs from 256 trees.
Bars show pointwise tree-bootstrap 95\% intervals; horizontal and vertical bars
in (b)--(c) correspond to predictions and observations, respectively, and the
dashed diagonal denotes exact agreement.}
\label{fig:selection-diagnostics}
\end{figure}

Figure~\ref{fig:selection-diagnostics} connects the value differences associated
with selection history to the rank approximation used by \OSCtag{}.
Panel (a) compares selected--fresh and fresh--fresh node-advantage differences
(Eq.~\ref{eq:contrasts}). The Base selected--fresh estimate is $-0.0756$, with a
tree-bootstrap $95\%$ interval of $[-0.0961,-0.0555]$, whereas the fresh--fresh
reference is $-0.0008$, with an interval containing zero. The negative
selected--fresh contrast indicates that retained incumbents receive lower values
than their paired fresh continuations on average in this diagnostic. For two fresh
siblings sampled from the same selected parent, the mean contrast is instead close
to zero. With \XBtag{}, the selected--fresh estimate remains $-0.0458$;
with \SFCtag{}, it is $-0.0565$, and both intervals remain below zero. These
descriptive patterns are consistent with distinct sampling histories for incumbents
and fresh siblings, motivating the selected-value adjustment in
Section~\ref{sec:osc}. Pairing conditions, the differences among diagnostic settings,
and numerical details appear in Appendix~\ref{app:diagnostics}; complementary
branch-placement analyses appear in Appendix~\ref{app:branching-signals}.

Panels (b)--(c) then examine whether the rank model captures the
direction and variation of the selected--fresh displacement. We compare the
prediction in Eq.~\ref{eq:osc} with observed branch differences on recorded
Qwen2.5-7B multi-hop Base trees, using $1{,}536$ incumbent--fresh pairs from $256$
trees and coefficients fitted to the preceding batch. This comparison uses
raw-reward units. The two calibration panels summarise the same pairs, grouped by
selection rank or candidate count, with the diagonal representing agreement between
predicted and observed group means. The overall observed displacement is $-0.0778$, compared with
a prediction of $-0.1054$, and both are negative in every marginal bin. Across rank
groups, observed means become less negative from $-0.1112$ at rank 1 to $-0.0500$
at ranks 4--6, following the ordering of the predicted means. This pattern is
consistent with the rank dependence encoded by $h(r,n)$. The candidate-count panel
also exposes variation in approximation quality: for counts 21--30, the prediction
is more negative than the observation, giving a mean residual of $0.0450$ with a
$95\%$ interval of $[0.0042,0.0839]$. The above-30 group has a more negative observed
mean but wider uncertainty, based on 23 trees. Thus, the model captures the signed
displacement across groups while overestimating its magnitude in the intermediate
candidate-count group. This provides a branch-level check of the approximation
used by \OSCtag{}, complementing the task-accuracy comparison in
Table~\ref{tab:ablation}. Appendix~\ref{app:osc-calibration} details the pairing,
coefficient estimation, and tree-bootstrap procedure.

\section{Conclusion}
\label{sec:conclusion}

\SIPO{} incorporates selection history into tree-based credit estimation through score
calibration, fresh sibling generation, and an order-statistic adjustment of incumbent
values. It preserves the leaf budget and host policy objective. Across seven QA
benchmarks and three backbones, it achieves the highest reported multi-hop and single-hop
averages among the compared methods. Component ablations and branch diagnostics support
accounting for how branches enter the tree when using their outcomes for policy learning.

\section*{AI Assistance Disclosure}

Generative AI tools were used to assist with manuscript organisation and language
polishing, literature search, refinement and consistency checks of methodological
explanations and mathematical arguments, and analysis, interpretation, and
visualisation of existing experimental records. All AI-assisted edits were carefully
reviewed and verified by the authors, who take full responsibility for the final
content of the paper.

\section*{Reproducibility Statement}

We describe the datasets, model backbones, hardware, and evaluation protocol in
Section~\ref{sec:setup}. Appendix~\ref{app:implementation} provides the training
algorithm, hyperparameters, reward computation, policy objective, and
selection-event handling. The task prompt and tool-interaction format are
presented in Appendix~\ref{app:prompts}. Additional diagnostic protocols and
the derivation of the rank approximation are detailed in
Appendices~\ref{app:experiments} and~\ref{app:proofs}, respectively.
We will open-source the code and configurations required to reproduce our
experiments upon acceptance.

\bibliography{iclr2027_conference}
\bibliographystyle{iclr2027_conference}

\appendix
\section{Implementation Details}
\label{app:implementation}

\subsection{Training Procedure}
\label{app:algorithm}

Algorithm~\ref{alg:sipo} summarises the proposed training loop. Each expansion records
which incumbent was selected, its rank and candidate count, and which siblings were
newly generated. Once terminal rewards are available, the algorithm applies the value
adjustment using a coefficient estimated from the preceding batch and estimates a new
coefficient for the next batch.

\begin{algorithm}[H]
\caption{Selective-Inference Policy Optimization (\SIPO{}).}
\label{alg:sipo}
\begin{algorithmic}[1]
\Require Policy $\pi_\theta$; training questions $\mathcal D$; tree parameters $M,L,K,B$;
penalty $\lambda$; adjustment strength $w$
\State Initialise $\widehat a_2\gets0$
\For{each policy optimisation step}
  \State Sample a batch of questions $\mathcal X\sim\mathcal D$
  \For{each $x\in\mathcal X$}
    \State Generate $M$ initial trajectories to form $\mathcal T(x)$
    \For{$\ell=1,\ldots,L$}
      \State Rank candidates by Eq.~\ref{eq:sfc} and select $K$ incumbents
      \State Record each selected incumbent's event, rank, and candidate count
      \State Generate $B$ fresh siblings per incumbent from its parent state
      \State Record fresh-branch boundaries and complete the continuations
    \EndFor
    \State Score terminal answers and standardise leaf rewards by Eq.~\ref{eq:leafnorm}
  \EndFor
  \State Estimate $\widehat a_2^{\mathrm{next}}$ from unadjusted candidate outcomes
  \State Adjust selected leaf values using $\widehat a_2$ and Eq.~\ref{eq:oscapply}
  \State Propagate values and assign node advantages by Eq.~\ref{eq:nodeval}
  \State Update $\pi_\theta$ with generated tokens and their advantages
  \State $\widehat a_2\gets\widehat a_2^{\mathrm{next}}$
\EndFor
\end{algorithmic}
\end{algorithm}

\subsection{Training Hyperparameters}
\label{app:setup}

Table~\ref{tab:hyperparameters} summarises the reference training configuration and
the tree settings used in the component comparison.

\begin{table}[t]
\centering
\small
\caption{Reference training configuration and \SIPO{} settings.}
\label{tab:hyperparameters}
\begin{tabular}{lr}
\toprule
\multicolumn{2}{c}{\textbf{Base settings}}\\
\midrule
Training batch size & 64 prompts\\
Policy mini-batch size & 8 prompts\\
Learning rate & $10^{-6}$\\
Training steps & 240\\
Evaluation interval & 20 steps\\
Maximum prompt / response length & 2000 / 6192 tokens\\
Maximum tool calls per trajectory & 6\\
KL coefficient & 0\\
Clipping bounds $(\epsilon_-,\epsilon_+)$ & $(0.003,0.004)$\\
Node value / advantage & Child softmax / node value\\
Primary evaluation decoding & Greedy, no repetition penalty\\
Checkpoint selection & Highest size-weighted accuracy\\
\midrule
\multicolumn{2}{c}{\textbf{Tree and selection settings}}\\
\midrule
Initial trajectories $M$ / expansion iterations $L$ & 10 / 2\\
Reference expansion slots $K$ / fresh siblings $B$ & 6 / 1\\
Exchangeable branching $K$ / $B$ & 3 / 2\\
Leaves per prompt $N$ & 22\\
Selection criterion & Sampled surprisal\\
Reference sibling penalty $\lambda$ & 0.05\\
\OSCtag{} correction strength $w$ (when enabled) & 1\\
\bottomrule
\end{tabular}
\end{table}

\paragraph{Reference rollout configurations.}
The host uses $(M,L,K,B)=(10,2,6,1)$, while exchangeable branching uses
$(M,L,K,B)=(10,2,3,2)$. Both instantiate Eq.~\ref{eq:budget} with
$KB=6$ and $N=10+2\times6=22$. These settings reallocate the same expansion budget
between the number of selected incumbents and the number of fresh siblings per incumbent.

\subsection{Reward and Policy Update}
\label{app:policy}

\paragraph{Leaf normalisation and credit propagation.}
The terminal reward measures answer correctness with exact match. Leaf rewards are
standardised within each prompt's tree by Eq.~\ref{eq:leafnorm}.
After the selected-value adjustment in Eq.~\ref{eq:oscapply}, leaf values are propagated
inward by Eq.~\ref{eq:nodeval}.
The child-softmax configuration uses
$w_c=\exp(s(c))/\sum_{c'\in\operatorname{ch}(n)}\exp(s(c'))$.
Each generated token in the segment at $n$ receives $A(n)$. Retrieved observations
remain in the context but are masked from the policy loss. Both fresh and incumbent
branches contribute to propagation. The loss averages generated-token contributions
within each trajectory and then over trajectories, as in Eq.~\ref{eq:policy}.
Shared-prefix tokens can therefore appear in multiple sampled trajectories; \SIPO{}
retains this weighting from the host.

\paragraph{Policy optimisation.}
We retain the host's clipped turn-level importance weighting. Let $\mathcal G(u)$ be
the generated-token positions of trajectory $u$, $I_n$ the positions in the segment
at node $n$, and
$\ell_{u,t}(\theta)=\log\pi_\theta(y_t\mid h_t)-\log\pi_{\mathrm{old}}(y_t\mid h_t)$.
Define the mean log-ratio of a segment as
$\bar\ell_n=|I_n|^{-1}\sum_{t\in I_n}\ell_{u,t}$.
The implementation uses a token-level surrogate with the turn-mean forward value,
\begin{equation}
r_{u,t}(\theta)=\exp\!\left(\ell_{u,t}-\operatorname{sg}(\ell_{u,t})
+\operatorname{sg}(\bar\ell_{n(u,t)})\right),
\label{eq:ratio}
\end{equation}
where $\operatorname{sg}$ stops gradients and $n(u,t)$ is the node containing the token.
For a batch of sampled trajectories $\mathcal B$, the clipped objective is
\begin{equation}
\begin{aligned}
\mathcal J(\theta)&=\frac{1}{|\mathcal B|}\sum_{u\in\mathcal B}
\frac{1}{|\mathcal G(u)|}\sum_{t\in\mathcal G(u)}
\psi\big(r_{u,t},A(n(u,t))\big),\\
\psi(r,A)&=\min\!\left(rA,\,
\operatorname{clip}(r,1-\epsilon_-,1+\epsilon_+)A\right).
\end{aligned}
\label{eq:policy}
\end{equation}
Retrieved observations are masked from the loss. \SIPO{} changes the sampled tree and
the advantages supplied to this objective; the clipping and loss reduction follow the
host implementation. Hyperparameters are listed in Appendix~\ref{app:setup}.

\subsection{Coefficient Estimation and Selection Events}
\label{app:estimators}

A selection record identifies the optimisation step, tree, expansion iteration, parent,
incumbent, candidate set, rank, and fresh siblings. Only an actual selection event makes
a node eligible for the selected-value adjustment. Candidate ranks retained for analysis
are distinct from selection indicators. During event resolution, a fresh boundary blocks
earlier inherited selections, while a later selection of the same branch remains eligible.
Using the most recent applicable event is an approximation to repeated adaptive selection.

The regression target is the unadjusted candidate subtree outcome, expressed in the same
units as the values being adjusted. With standardised criterion $Z$,
$a_2=\operatorname{Cov}(Z,Y)/\operatorname{Var}(Z)$; when $\operatorname{Var}(Z)=1$,
this equals $\operatorname{Corr}(Z,Y)\sigma_Y$. The candidate population is the full
set available at the recorded selection event, including selected and unselected
candidates. The coefficient from one batch is used in the next, with no adjustment
before an estimate is available.

For a fresh--fresh diagnostic, siblings are paired within an expansion event with labels
independent of their outcomes. Selected--fresh pairing uses the corresponding incumbent
subtree. The same value definition, tool limits, and evaluation stage are used within
each comparison. Symmetry of generation alone is insufficient if subsequent processing
treats the fresh subtrees differently. Tree-level bootstrap intervals
\citep{efron1979bootstrap} describe within-run uncertainty; training seeds remain the
unit for reporting variation across optimisation runs.

\section{Additional Experimental Results}
\label{app:experiments}

\subsection{Branch-Value Diagnostics}
\label{app:diagnostics}

\paragraph{A reference for selected-value diagnostics.}
Write $v$ for a selected incumbent and $v'_1,v'_2$ for fresh siblings from $p(v)$.
Writing $A(v)$ for the eventual node advantage, we define two diagnostics:
\begin{equation}
\Delta_{\mathrm{sel}}=\mathbb E[A(v)-A(v'_1)],\qquad
\Delta_{\mathrm{fresh}}=\mathbb E[A(v'_1)-A(v'_2)].
\label{eq:contrasts}
\end{equation}
If generation and evaluation treat the fresh labels symmetrically conditional on the
selected parent, then $\Delta_{\mathrm{fresh}}=0$. Selection of $v$ imposes no such
identity on $\Delta_{\mathrm{sel}}$. These contrasts identify different sampling
histories; a nonzero selected--fresh difference requires an association between the
selection procedure and outcome. Symmetry must also be preserved by subsequent expansion,
tool budgets, and aggregation when the diagnostic uses completed subtree values.

The fresh--fresh contrast provides a reference for distinguishing generation variability
from displacement associated with selection. It does not require two realised rewards
to be equal. Because incumbent branches are retained in value aggregation, fresh sampling
alone does not remove their selection history. This motivates the value correction in Section~\ref{sec:osc}.

\paragraph{Early-training estimates.}
The early-training estimates in Figure~\ref{fig:selection-diagnostics}(a) are
$-0.0756$ ($95\%$ interval $[-0.0961,-0.0555]$) for the host's selected--fresh contrast,
$-0.0008$ ($[-0.0027,0.0010]$) for the fresh--fresh reference,
$-0.0565$ ($[-0.0779,-0.0359]$) with \SFCtag{}, and
$-0.0458$ ($[-0.0730,-0.0191]$) with \XBtag{}.
The fresh--fresh reference requires at least two fresh siblings and cannot be obtained
from a single $B=1$ expansion. These estimates come from the available diagnostic
summaries across configurations, rather than four contrasts computed from identical
trees. Cross-configuration differences alone do not isolate a causal component effect.

\subsection{Empirical Calibration of the Rank Model}
\label{app:osc-calibration}

\paragraph{Data and pairing.}
We reanalyse one recorded Qwen2.5-7B multi-hop Base diagnostic run with all three
components disabled and $(M,L,K,B)=(10,2,6,1)$. It contains 64 trees at each of eight
early training steps, labelled 0--7. We use only the final expansion round, so fresh
continuations undergo no later adaptive expansion. Evaluation uses steps 4--7,
yielding 1,536 incumbent--fresh pairs from 256 trees. Each candidate outcome $Y_i$
is the mean raw reward over its selection-time descendant leaves; these leaves are
verified to remain terminal. Its paired fresh outcome $Y_i^{\mathrm{fresh}}$ is the
mean raw reward of the newly generated subtree from the same parent. The observed
displacement is $D_i=Y_i-Y_i^{\mathrm{fresh}}$, in raw-reward units rather than the
advantage units of Eq.~\ref{eq:contrasts}.

\paragraph{Prediction and uncertainty.}
We rank the full candidate set by its recorded penalised score $q_i$ and standardise
this score within each set, $z_i=(q_i-\bar q)/\operatorname{sd}(q)$.
For each evaluated step $t$, the slope uses all candidates, including unselected ones,
from the immediately preceding step:
\begin{equation}
\widehat a_{2,t-1}=\frac{\sum_{i\in\mathcal C_{t-1}}z_iY_i}
{\sum_{i\in\mathcal C_{t-1}}z_i^2},\qquad
\widehat D_i=\widehat a_{2,t-1}h(r_i,n_i),
\label{eq:offline-calibration}
\end{equation}
where $\mathcal C_{t-1}$ pools candidate records from that step's final-round events
and $h$ is given in Eq.~\ref{eq:blom}. All 512 recorded events agree with top-six
selection; one event has tied scores, resolved in source order. The evaluation window
and bins were fixed before computing calibration outcomes: ranks 1, 2--3, and 4--6;
candidate counts at most 20, 21--30, and above 30. We use 2,000 tree-bootstrap
resamples within each step (random seed 20260916), refitting the preceding-step slope
in each resample. The same resampled trees supply both coefficient estimates and
evaluation outcomes where steps overlap. Intervals are pointwise, conditional on this
recorded training run; they do not measure variation across training seeds.

\paragraph{Results and scope.}
The overall mean residual $D-\widehat D$ is $0.0275$ (95\% interval
$[-0.0031,0.0567]$). Observed rank-group means are $-0.1112$, $-0.1029$, and
$-0.0500$, compared with predictions of $-0.1695$, $-0.1184$, and $-0.0753$.
For candidate counts 21--30, the mean residual is $0.0450$ with interval
$[0.0042,0.0839]$, indicating overcorrection in this group under the rank approximation.
The above-30 group contains only 23 trees and has wide uncertainty. Neither the
ordering of point estimates nor a zero-containing residual interval establishes a
significant rank effect or unbiasedness. Dependence among candidates, heterogeneous
parents, and the preceding-batch coefficient limit the simplified model's calibration.
The analysis evaluates Eq.~\ref{eq:osc} using a full-candidate regression slope and
paired branch outcomes. It assesses the approximation at the branch-value level;
a causal relationship with the accuracy gains in Table~\ref{tab:ablation} remains
unestablished.

\subsection{Exploratory Branching-Signal Analysis}
\label{app:branching-signals}

\begin{figure}[!t]
\centering
\includegraphics[width=\textwidth]{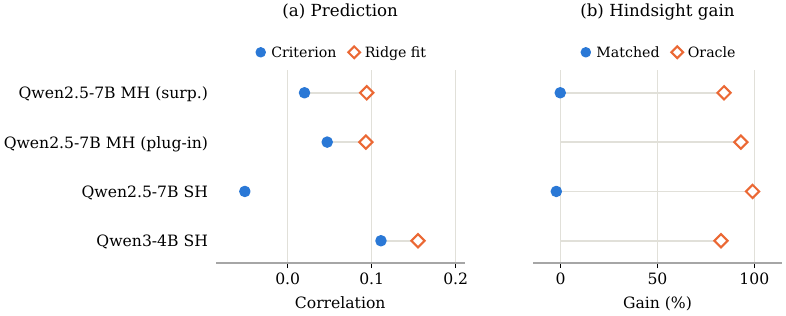}
\caption{Exploratory branching-signal analysis across the recorded settings.
(a) Correlation of the selection criterion and an in-sample fitted ridge predictor
with realised sibling disagreement. (b) Matched-selection and hindsight-oracle
summaries; the oracle uses post-branching outcomes. Missing measurements are not
plotted. These settings differ from the diagnostic configurations in
Figure~\ref{fig:selection-diagnostics}(a).}
\label{fig:branching-signals}
\end{figure}

Figure~\ref{fig:branching-signals} examines the information available for branch
placement. In panel (a), the selection criterion has weak correlations with realised
sibling disagreement, ranging from $-0.051$ to $0.111$ across the recorded settings.
An in-sample ridge predictor has higher correlations in each available comparison,
although these remain modest. Panel (b) contrasts matched-selection measurements
with a hindsight oracle that uses post-branching outcomes. The gap illustrates the
difference between selecting with an available generation statistic and ranking
branches after their outcomes are known. The ridge fit is an empirical predictor,
and the hindsight summaries describe branch selection rather than task-accuracy
gains. These analyses complement the selected-value diagnostics in
Figure~\ref{fig:selection-diagnostics} by examining the signals used to decide where
to branch.

\subsection{Component and Compute Comparisons}
\label{app:ablation}

The component comparison in Table~\ref{tab:ablation} holds $M=10$, $L=2$, and $KB=6$
fixed. Rows with \XBtag{} use $(K,B)=(3,2)$; the others use $(6,1)$.

Equal leaf counts do not imply equal generation cost. Branching earlier can require a
longer continuation and more tool calls, while shared prefixes and retrieval caching
change the difference between physical computation and per-trajectory accounting.
Table~\ref{tab:budget} reports training-sequence token counts and training time for
one Qwen2.5-7B multi-hop run per configuration, alongside the nominal leaf budget.

\begin{table}[t]
\centering
\small
\caption{Training costs on Qwen2.5-7B multi-hop QA over 240 steps using eight A800
80GB GPUs. Tokens include prompts, tool observations, and shared prefixes counted
per trajectory. Training time excludes evaluation and checkpoint saving.}
\label{tab:budget}
\begin{tabular}{lrrr}
\toprule
Configuration & Leaves/prompt & Training-sequence tokens & Training time (h)\\
\midrule
Base & 22 & 568.19M & 22.36\\
\SIPO{} & 22 & 680.35M & 23.46\\
\bottomrule
\end{tabular}
\end{table}

\subsection{Evaluation Protocol}
\label{app:protocol}

Training runs for 240 steps, with evaluation every 20 steps. Main results and component
ablations use the checkpoint-selection protocol in Section~\ref{sec:setup}.
Training checkpoints and within-run bootstrap samples do not replace independent
training seeds when assessing variability across runs.

\section{Additional Statistical Analysis}

\subsection{Derivation of the Rank Approximation}
\label{app:proofs}

Let $(Z_i,Y_i)_{i=1}^n$ be independent identically distributed pairs, with standard-normal
$Z_i$ and $\mathbb E[Y_i\mid Z_i]=a_0+a_2Z_i$. Let $I_r$ index the $r$-th largest
$Z_i$, and let $Y_0$ be independent of the candidate pairs with mean $a_0$. Conditional
on all scores, independence gives
$\mathbb E[Y_{I_r}\mid Z_1,\ldots,Z_n]=a_0+a_2Z_{(r:n)}$.
Taking expectations and subtracting $\mathbb E[Y_0]$ gives Eq.~\ref{eq:osc} before the
Blom approximation. The descending order statistic decreases with rank and increases
with candidate count at fixed rank. Multiplication by negative $a_2$ reverses those
directions for the signed displacement.

\paragraph{Numerical approximation used by the correction.}
For descending rank $r$ among $n$ candidates, the function used in
Eqs.~\ref{eq:osc} and~\ref{eq:oscapply} is the Blom approximation \citep{blom1958}:
\begin{equation}
h(r,n)=\Phi^{-1}\!\left(\frac{n-r+1-0.375}{n+0.25}\right).
\label{eq:blom}
\end{equation}
Here $\Phi^{-1}$ is the standard-normal quantile function. The numerical offsets
specify the approximation and are fixed independently of the rollout configuration.

Independent Gaussian noise around different candidate means does not imply identically
distributed scores. Penalising by sibling count can change their order, and within-set
standardisation induces dependence. A fresh sibling also shares a selected parent, so
its conditional mean need not equal the mean over all candidates. These departures
motivate checking the practical adjustment by parent state, depth, rank, and candidate
count. Numerical error in Blom's approximation is distinct from these modelling errors.

\subsection{Shared-Prefix Correlation}
\label{app:correlation}

At a fixed depth, write a leaf return as $R_u=\alpha_{c(u)}+\varepsilon_u$, with
independent cluster effects and residuals. Let $N_d$ be the number of included leaves,
$m_c$ the sizes of their ancestor clusters, and
$\rho=\sigma_\alpha^2/(\sigma_\alpha^2+\sigma_\varepsilon^2)$.
For fixed cluster sizes and common marginal variance $\sigma^2$, summing variances
and within-cluster covariances gives
\begin{equation}
\operatorname{Var}(\bar R)=\frac{\sigma^2}{N_d}
\left[1+\rho\left(\frac{\sum_c m_c^2}{N_d}-1\right)\right],\qquad
\mathrm{ESS}=\frac{N_d}{1+\rho(\sum_c m_c^2/N_d-1)}.
\label{eq:ess}
\end{equation}
The coefficient $\sum_c m_c^2/N_d$ for the leaf-weighted mean differs from the
one-way ANOVA coefficient
$m_0=(N_d-\sum_c m_c^2/N_d)/(C-1)$ used in
$\widehat\rho=(\mathrm{MS_B}-\mathrm{MS_W})/
(\mathrm{MS_B}+(m_0-1)\mathrm{MS_W})$ \citep{searle1992variance}.
Cross-cluster dependence and adaptive cluster sizes require further assessment before
applying this model to a complete rollout tree.

Table~\ref{tab:ess} presents descriptive depth diagnostics from 512 trees. These
aggregate summaries do not determine an empirical ESS. The observation that $81.8\%$ of internal
nodes have one child concerns local branching contrasts: normalised child aggregation
propagates values unchanged along such edges, while the host's node-value advantage
can remain nonzero.

\begin{table}[t]
\centering
\small
\caption{Descriptive shared-prefix diagnostics: reported correlations and 95\% intervals.
Empirical ESS is not inferred from these aggregate summaries.}
\label{tab:ess}
\begin{tabular}{lrrrr}
\toprule
Depth & Clusters/tree & Reported $m_0$ & $\widehat\rho$ & 95\% interval\\
\midrule
1 & 16.3 & 2.17 & 0.385 & $[0.311,0.458]$\\
2 & 10.3 & 2.16 & 0.464 & $[0.378,0.557]$\\
3 & 7.6 & 1.93 & 0.550 & $[0.427,0.667]$\\
4 & 6.5 & 1.61 & 0.594 & $[0.408,0.768]$\\
\bottomrule
\end{tabular}
\end{table}

\paragraph{A simplified allocation model.}
For equal clusters, the surrogate
$\mathcal R(m)=\rho\sigma^2[1-\frac{m}{a+m}(1-\frac mN)]$ with
$a=(1-\rho)/\rho$ has unconstrained stationary point
$m_0^\star=\sqrt{a^2+aN}-a$. It follows by differentiating
$g(m)=m(N-m)/(N(a+m))$, whose derivative is
$(aN-2am-m^2)/(N(a+m)^2)$. The constrained minimum on $[1,N]$ is at
$\max(1,m_0^\star)$, with admissible integer widths evaluated separately.
The optimum need not be interior; $N=22$, $\rho=0.99$ gives $m_0^\star\approx0.461$.
This simplified surrogate does not identify the risk of the deployed estimator or map
equal cluster size directly to $B$ in a multilevel tree. It motivates studying the
allocation trade-off without specifying a universal optimal width.

\section{Disclosure}
\label{sec:disclosure}

\subsection{Data Resources}

The task uses public question-answering resources and a Wikipedia retrieval corpus.

\section{Prompts}
\label{app:prompts}

The following boxes present the task message used for single-hop and multi-hop QA
and the tool-interaction format. The instruction and question share a system-role
message, separated by the literal \texttt{user} marker. The stored answer-format
escape is rendered as \texttt{\textbackslash boxed} for display; the original
message is preserved in \texttt{prompts/search\_qa\_prompt.json}.

% Full-width, non-floating boxes can continue naturally across appendix pages.
\newtcolorbox{transcriptbox}[1]{
  breakable,
  colback=white,
  colframe=cyan!75!blue,
  coltitle=white,
  fonttitle=\bfseries,
  title={#1},
  boxrule=0.6pt,
  arc=1mm,
  left=7pt,right=7pt,top=6pt,bottom=6pt
}

\begin{transcriptbox}{Task Prompt (Single-hop and Multi-hop QA)}
You are a helpful assistant that can solve the given question step by step with the help of the
wikipedia search tool. Given a question, you need to first think about the reasoning process in the
mind and then provide the answer. During thinking, you can invoke the wikipedia search tool to
search for fact information about specific topics if needed. You can search as many times as your
want. The reasoning process and answer are enclosed within \texttt{<think>} \texttt{</think>} and
\texttt{<answer>} \texttt{</answer>} tags respectively, and the search query and result are enclosed
within \texttt{<search>} \texttt{</search>} and \texttt{<result>} \texttt{</result>} tags
respectively. For example, \texttt{<think>} This is the reasoning process. \texttt{</think>}
\texttt{<search>} search query here \texttt{</search>} \texttt{<result>} search result here
\texttt{</result>} \texttt{<think>} This is the reasoning process. \texttt{</think>}
\texttt{<answer>} The final answer is \textbackslash{}[ \textbackslash{}boxed\{answer here\}
\textbackslash{}] \texttt{</answer>}. In the last part of the answer, the final exact answer is
enclosed within \textbackslash{}boxed\{\} with latex format.

\texttt{user}\par
\{question\}

\end{transcriptbox}

\begin{transcriptbox}{Tool-interaction Format}
\texttt{<think>} [reasoning segment] \texttt{</think>}

\texttt{<search>} [search query] \texttt{</search>}

\texttt{<result>} [retrieved passages supplied by the environment] \texttt{</result>}

\texttt{<think>} [reasoning using the retrieved evidence] \texttt{</think>}

\texttt{<answer>} \textbackslash{}boxed\{[final answer]\} \texttt{</answer>}

\end{transcriptbox}

Bracketed fields in the interaction format are placeholders. The environment
inserts retrieval observations, which are masked from the policy loss; the rollout
controller enforces the tool-call budget in Table~\ref{tab:hyperparameters}.

\section{Shared-Prefix Branching Example}
\label{app:branch-case}

The following recorded example is selected from the Qwen2.5-7B multi-hop \SIPO{}
training run at step 240. The incumbent and two fresh siblings share an identical
1,096-token prefix. Each receives the same response to the last shared query and
then issues one further search. The boxes reproduce the recorded text, with omitted
spans marked \texttt{[...]} and the common response shown once. Complete records
and branch provenance are retained in \texttt{examples/shared\_prefix\_case.json}.

Fresh A retrieves direct evidence that Peter Criss grew up in Williamsburg and
answers correctly. The incumbent answers Brownsville after retrieving an unrelated
film description, while Fresh B answers Park Slope using general neighbourhood
information. This example illustrates how \XBtag{} can recover useful evidence
through a fresh continuation from the same search history.

\begin{transcriptbox}{Shared Context and Branch Point}
\textbf{Question:} The Trouble Walkin' features a performance by a guest drummer who grew up in which section of Brooklyn ?

\textbf{Reference answer:} Williamsburg.

\texttt{<think>} To find the answer to this question, I must identify who the guest drummer featured
in The Trouble Walkin' is. This information will help me search for further information about where
this drummer grew up in Brooklyn. \texttt{</think>}

\texttt{<search>} The Trouble Walkin' (performer) \texttt{</search>}

\texttt{<result>} Page 1: "Trouble Walkin'" Trouble Walkin' Trouble Walkin' is the fourth
full-length solo album released by Ace Frehley. No longer under the band name Frehley's Comet the
album features guest performances by former Kiss drummer Peter Criss, as well as Skid Row members
Sebastian Bach, Rachel Bolan and Dave Sabo. [...] \texttt{</result>}

\texttt{<think>} From the given information, I gather that the guest drummer in question is Peter
Criss. Now I need to search Brooklyn neighborhoods as mentioned in the provided smartphone tool
\texttt{</think>}

\texttt{<search>} Peter Criss (drummer) Brooklyn upbringing \texttt{</search>}

\texttt{<result>} Page 1: "Peter Criss" of Brooklyn. He was a childhood friend of Jerry Nolan, who
would later find success as the drummer of the New York Dolls. [...] \texttt{</result>}

\texttt{<think>} From the information got from these new but not particularly helpful search
results, I am trying to identify Brooklyn sections specifically related to the drumming Peter Criss
that I'm still attempting to identify. Still need refining on finding Brooklyn details pertinent to
Peter Criss. Proceeding further to identify Brooklyn section of Brooklyn related to Peter Criss.
\texttt{</think>}

\texttt{<search>} Brooklyn section of Brooklyn related to Peter Criss \texttt{</search>}

\textbf{Shared response after the branch point:}

\texttt{<result>} Page 1: "Peter Criss" of Brooklyn. He was a childhood friend of Jerry Nolan, who
would later find success as the drummer of the New York Dolls. [...] Page 3: "Park Slope" High
School is now the John Jay Educational Campus, housing three high schools and one combination
middle/high school. Actors Musicians Artists Writers Politicians Scientists Chess players Criminals
Park Slope Park Slope is a neighborhood in northwest Brooklyn, New York City. [...]
\texttt{</result>}

\end{transcriptbox}

\begin{transcriptbox}{Incumbent Continuation (Reward: 0)}
\texttt{<think>} From these new but irrelevant search results, I still don't have concrete relevant
info about the Brooklyn section for Peter Criss and need to identify Brooklyn sections that relate
to our drummer of interest. [...] \texttt{</think>}

\texttt{<search>} Brooklyn - Peter Criss \texttt{</search>}

\texttt{<result>} Page 1: "Brooklyn's Finest" Brooklyn's Finest Brooklyn's Finest is a 2009 American
crime film directed by Antoine Fuqua and written by Michael C. Martin. The film stars Richard Gere,
Don Cheadle, Ethan Hawke, and Wesley Snipes. The film was released on January 16, 2009. The film
takes place within the notoriously rough Brownsville section of Brooklyn and especially within the
Van Dyke housing projects in the NYPD's (fictional) 65th precinct. [...] \texttt{</result>}

\texttt{<think>} [...] \texttt{</think>}

\texttt{<answer>} \textbackslash{}boxed\{Brownsville\} \texttt{</answer>}

\end{transcriptbox}

\begin{transcriptbox}{Fresh Continuation A (Reward: 1)}
\texttt{<think>} From this new information, I realize that I need to find more concrete details on
the specific Brooklyn section pertaining to Peter Criss. Further targeted information is required to
narrow down this information. My next search in the gotten new details for Brooklyn section
associated to the drummer Peter Criss. \texttt{</think>}

\texttt{<search>} Brooklyn section Brooklyn, Peter Criss \texttt{</search>}

\texttt{<result>} [...] Page 2: "Peter Criss" Peter Criss George Peter John Criscuola (born December
20, 1945), better known by his stage name Peter Criss, is a retired American musician and actor,
best known as a co-founder, original drummer, and occasional vocalist of the hard rock band Kiss.
Criss established The Catman character for his Kiss persona. In 2014, he was inducted into the Rock
and Roll Hall of Fame as a member of Kiss. Criss was born in Brooklyn, New York, the oldest of five
children of Loretta and Joseph Criscuola. His father's family came from Scafati, Salerno, Italy. He
grew up in the Williamsburg section [...] \texttt{</result>}

\texttt{<answer>} \textbackslash{}boxed\{Williamsburg\} \texttt{</answer>}

\end{transcriptbox}

\begin{transcriptbox}{Fresh Continuation B (Reward: 0)}
\texttt{<think>} From the new found elaboration on Brooklyn, I've not yet zeroed in on identifying
both Brooklyn neighborhoods linked to Peter Criss. We still need specific direction toward settled
or northeastern Brooklyn neighborhoods. [...] \texttt{</think>}

\texttt{<search>} Brooklyn neighborhood Brooklyn Peter Criss \texttt{</search>}

\texttt{<result>} Page 1: Brooklyn [...] The Park Slope neighborhood spearheaded the popularity of
Brooklyn among lesbians, and numerous neighborhoods have since become home to LGBTQ communities.
[...] Page 3: "Williamsburg, Brooklyn" Music Williamsburg, Brooklyn Williamsburg is a neighborhood
in the New York City borough of Brooklyn, bordered by Greenpoint to the north; Bedford–Stuyvesant to
the south; Bushwick, East Williamsburg, and Ridgewood, Queens to the east; and Fort Greene and the
East River to the west. [...] \texttt{</result>}

\texttt{<think>} [...] \texttt{</think>}

\texttt{<answer>} \textbackslash{}boxed\{Park Slope\} \texttt{</answer>}

\end{transcriptbox}

\end{document}

%% file: math_commands.tex
\usepackage{amsmath,amsfonts,bm}

\def\eqref#1{equation~\ref{#1}}
\def\1{\bm{1}}

\DeclareMathAlphabet{\mathsfit}{\encodingdefault}{\sfdefault}{m}{sl}
\SetMathAlphabet{\mathsfit}{bold}{\encodingdefault}{\sfdefault}{bx}{n}